\documentclass[sigconf,screen]{acmart}

\AtBeginDocument{%
  \providecommand\BibTeX{{%
    \normalfont B\kern-0.5em{\scshape i\kern-0.25em b}\kern-0.8em\TeX}}}

\copyrightyear{2023} 
\acmYear{2023} 
\setcopyright{rightsretained} 
\acmConference[WWW '23 Companion]{Companion Proceedings of the ACM Web Conference 2023}{April 30-May 4, 2023}{Austin, TX, USA}
\acmBooktitle{Companion Proceedings of the ACM Web Conference 2023 (WWW '23 Companion), April 30-May 4, 2023, Austin, TX, USA}
\acmDOI{10.1145/3543873.3587308}
\acmISBN{978-1-4503-9419-2/23/04}

\usepackage[ruled,lined,boxed,commentsnumbered]{algorithm2e}
\usepackage{microtype}
\usepackage{multirow}
\usepackage{multicol}
\usepackage{amsmath}
\usepackage{booktabs,subcaption,amsfonts,dcolumn}
\usepackage{color} 
\usepackage{balance} 
\usepackage{url}
\usepackage[utf8]{inputenc}
\usepackage{graphicx}

\SetKwRepeat{Do}{do}{while}%

\usepackage{CJKutf8}
\usepackage{enumitem}
\begin{document}

\title{A Concept Knowledge Graph for User Next Intent Prediction at Alipay}


\author{Yacheng He}
\affiliation{%
  \institution{Ant Group}
  \city{Hangzhou}
  \country{China}
}
\email{heyachen.hyc@antgroup.com}

\author{Qianghuai Jia}
\authornote{Corresponding author.}
\affiliation{%
  \institution{Ant Group}
  \city{Hangzhou}
  \country{China}
}
\email{qianghuai.jqh@antgroup.com}

\author{Lin Yuan}
\affiliation{%
  \institution{Ant Group}
  \city{Hangzhou}
  \country{China}
}
\email{huiwai.yl@antgroup.com}

\author{Ruopeng Li}
\affiliation{%
  \institution{Ant Group}
  \city{Hangzhou}
  \country{China}
}
\email{ruopeng.lrp@antgroup.com}

\author{Yixin Ou}
\affiliation{%
  \institution{Zhejiang University}
  \city{Hangzhou}
  \country{China}
}
\email{ouyixin@zju.edu.cn}

\author{Ningyu Zhang}
\affiliation{%
  \institution{Zhejiang University}
  \city{Hangzhou}
  \country{China}
}
\email{zhangningyu@zju.edu.cn}

\renewcommand{\shortauthors}{Yacheng He, et al.}

\begin{CJK}{UTF8}{gbsn}

\begin{abstract}

This paper illustrates the technologies of user next intent prediction with a concept knowledge graph. The system has been deployed on the Web at Alipay\footnote{\url{https://global.alipay.com/platform/site/ihome}}, serving more than 100 million daily active users. To explicitly characterize user intent, we propose \textbf{AlipayKG}, which is an offline concept knowledge graph in the Life-Service domain modeling the historical behaviors of users, the rich content interacted by users and the relations between them. We further introduce a Transformer-based model which integrates expert rules from the knowledge graph to infer the online user's next intent. Experimental results demonstrate that the proposed system can effectively enhance the performance of the downstream tasks while retaining explainability.  

\end{abstract}



\maketitle

\section{Introduction}

\begin{figure}[tbh]
\centering
\includegraphics[width=1\linewidth]{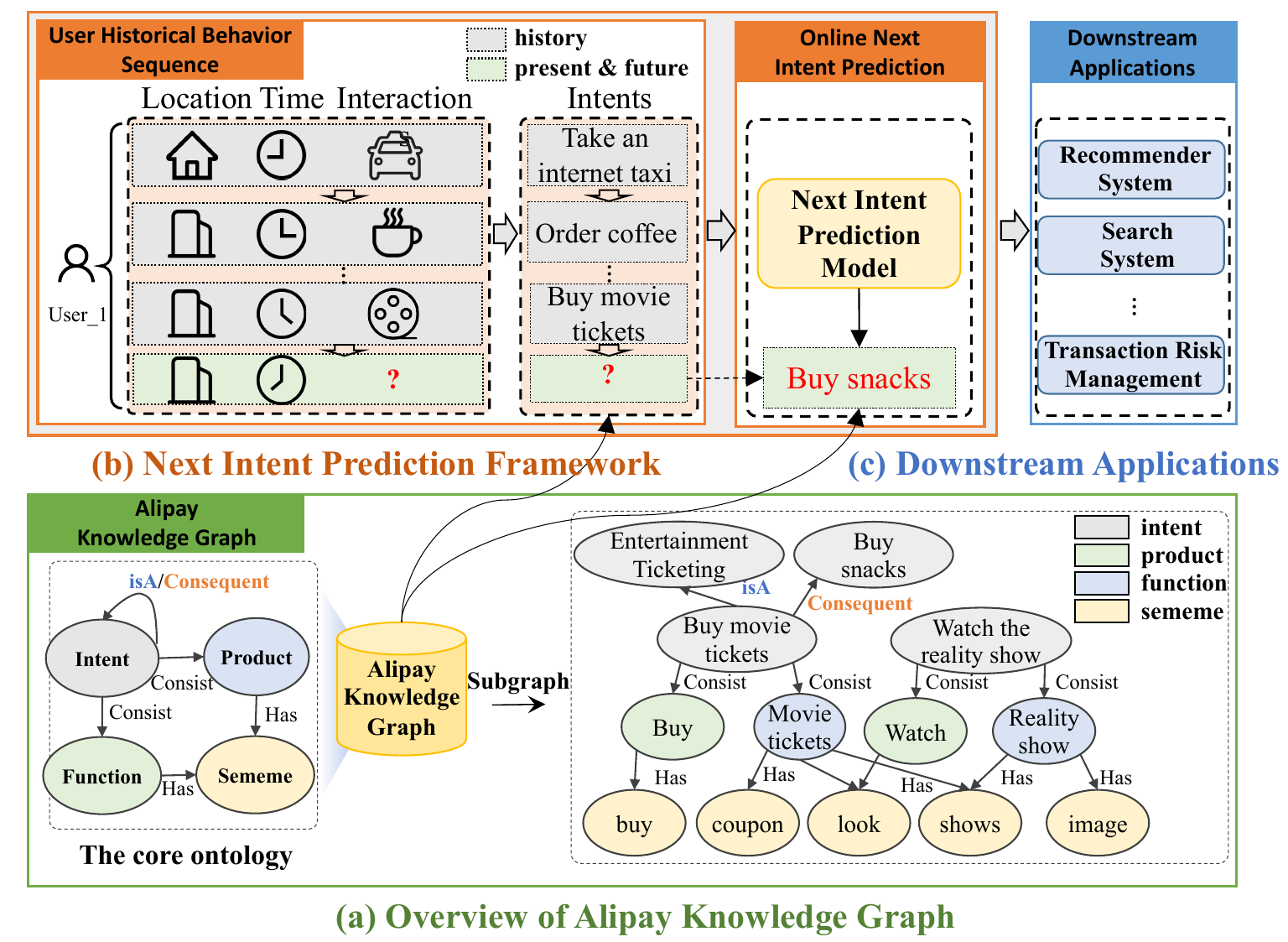}
 \caption{
The user next intent prediction system at Alipay. 
 Sub-figure (a) illustrates the core ontology and subgraph of AlipayKG. 
 In sub-figure (b), an example of the user's historical interactions and intent sequence is shown in gray-grounded boxes, and the next intent is marked with a red "?" that has been inferred as "buy snacks" by the next intent prediction model, whose outputs will provide a clear signal to downstream applications as shown in sub-figure (c).}
 \label{domain knowledge graph}
\end{figure}

User next intent prediction – the ability to automatically infer the next decision of users based on historical behavior and background knowledge – holds an important place in in-device Apps \cite{qu2019user}.
For example, in digital life service platforms such as Alipay, users often purchase snacks at the cinema (corresponding intent "buy snacks") after buying movie tickets via TaoPiaoPiao\footnote{\url{https://dianying.taobao.com/}} (corresponding intent "buy movie tickets"), which implies the intent of "buy movie tickets" may lead to the following intent of "buy snacks." 
As shown in Figure \ref{domain knowledge graph}, the ability to infer the future intents of users has the potential to be advantageous for tasks such as recommendation, searching, transaction risk management and so on.

Intuitively, user intent can be characterized as clustering patterns of user behaviors, which are usually hidden in the content and interacted with or generated by users in mobile applications. 
Specifically, the core of understanding user intent in Alipay lies in systematic and explicit knowledge modeling of the user's situation and the user's interacted item content, which consists of queries, applet services, bills, coupons, stores, reviews, etc.
Concretely, we summarize two non-trivial issues in user next intent prediction at Alipay as follows: 

\begin{itemize}[leftmargin=*]
\item \textbf{How to characterize user intent.} 
It is challenging to abstract and encode intent from user behaviors that are very diverse and can not be directly observed. 
In particular, unlike e-commerce scenarios such as Amazon\footnote{\url{https://www.amazon.com/}} which mainly contains shopping intent, the behaviors at Alipay are various, including shopping, trip and payment, which further increases the difficulty of intent representation.

\item \textbf{How to predict the user's next intent in real-time.} 
The user's next intent is not only based on the user's profile and preference but also largely influenced by spatial and temporal factors.
For example, the intent of "buy movie tickets" tends to occur at the weekend, while the intent of "order coffee" often occurs at Starbucks in the afternoon.

\end{itemize}

To address the above-mentioned issues, we propose a user next intent prediction system based on the Knowledge Graph (KG) and apply it to downstream applications at Alipay. 
We summarize the contributions of this paper as follows:

\begin{itemize}[leftmargin=*]
\item[$\bullet$] We propose \textbf{AlipayKG}, a concept knowledge graph that explicitly represents user behaviors by defining an intent architecture to achieve a unified representation of multi-source heterogeneous data. 
Meanwhile, we propose a systematic approach to obtain structured knowledge from those multi-source data. 
With the proposed AlipayKG, we address the first issue.
\item[$\bullet$] As for the second issue, we design a next intent prediction framework that integrates expert rules from AlipayKG, which improves the performance while increasing interpretability by expert rules during inference.
\item[$\bullet$] We evaluate this system on downstream tasks. 
Experimental results demonstrate that the proposed system deployed at Alipay can enhance the performance of several real-world applications, which serve more than 100 million daily active users.
\end{itemize}

\section{AlipayKG-Based User Next Intent Prediction System}

An overview of our user intent system is presented in Figure \ref{domain knowledge graph}, and it is composed of two parts as follows:
1) \textbf{AlipayKG} to sufficiently characterize user intent, 
and 2) \textbf{Next Intent Prediction Framework} to accurately predict the user’s next intent in real-time. 
\textbf{All collected data are anonymized and reviewed by the IRB committee at Alipay to preserve privacy.}

\subsection{AlipayKG Construction}
In general, user intent plays a crucial role in promoting the performance and interpretability of user modeling systems. 
However, uniformly capturing the users' intents and expressing them is arduous due to the various kinds of users' behaviors in digital life service platforms.
Therefore, to sufficiently characterize user intent, we propose a concept KG in the Life-Service domain called AlipayKG. 

\begin{figure}[t]
\centering
\includegraphics[width=1\linewidth]{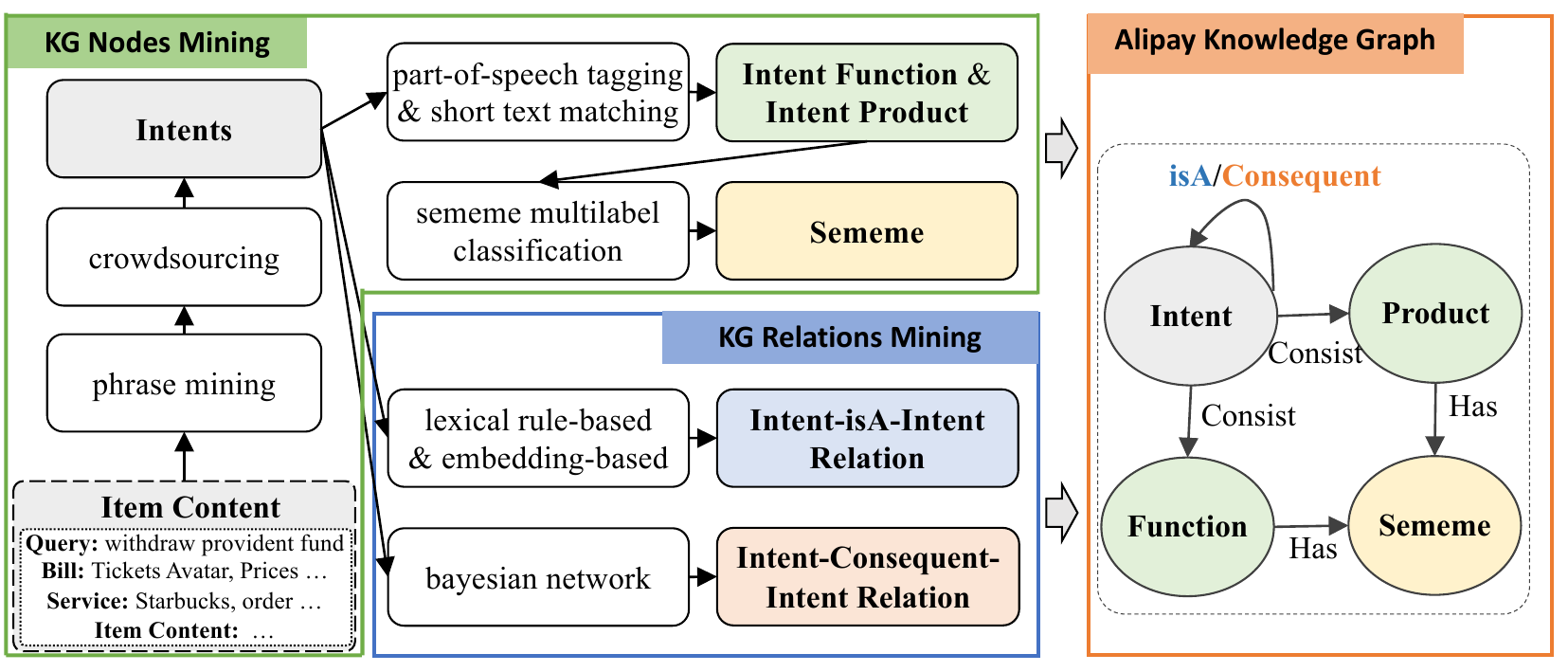}
\caption{
The process of constructing AlipayKG consists of nodes mining and relations mining (by different colors). 
}
\label{Knowledge mining process}
\end{figure}

The core ontology of AlipayKG is shown in Figure \ref{domain knowledge graph}(a), which includes four types of nodes and four types of relations. 
Specifically, "Intent" describes the decision drivers behind users' needs and mainly consists of "Function" and "Product," such as "take an internet taxi 打网约车" and "order coffee 点咖啡." 
Furthermore, "Product" and "Function" can be represented by more fine-grained Hownet sememes\footnote{\url{https://github.com/thunlp/OpenHowNet.git}} \cite{HowNet} that are regarded as the basic units for semantic understanding, such as "movie ticket|电影票 = \{coupon|票证, look|看, shows|表演物\}." 
Meanwhile, we introduce two specific types of relations between "Intent" nodes: 
1) "isA" relation builds the semantic hyponymy of "Intent" nodes, such as "rent an iPhone13 -isA- rent a mobile phone"; 
2) "Consequent" relation is used to establish the order effect of "Intent" nodes, such as "buy a house -consequent- renovate a house." 
Figure \ref{Knowledge mining process} illustrates the framework of AlipayKG construction, which contains two parts: 1) KG Nodes Mining and 2) KG Relations Mining. 
It is worth noting that crowdsourcing (we train hundreds of human annotators and ask them to filter incorrect facts in KG) is employed for data quality control throughout the whole process. \par

\subsubsection{KG Nodes Mining}
To mine "Intent" nodes, we adopt the automated phrase mining approach \cite{AliMeKG} based on item content and extend it with a self-constructed ground dictionary for high-quality phrase classification, where item content is chosen as our data source since users often directly express their requirements by interacting with items. 
Although the items in Alipay are multi-source heterogeneous, the text of different items shares the same critical information and can be used as input data sources for knowledge mining. 
Then, we utilize lexical rule matching, part-of-speech tagging, and short text matching models and structure the "Intent" nodes into two parts: "Function" and "Product."
Moreover, HowNet \cite{HowNet} has a wealth of artificially annotated corpus, through which we train a multi-label classification model \cite{Asymmetric} to automatically obtain sememe information of "Function" and "Product."
Due to aliasing and ambiguity issues with node names, we further use alignment models of Bert-Int \cite{BERT-INT} on "Intent" and "Product" nodes for semantic disambiguation, respectively. \par

\subsubsection{KG Relations Mining}
In this part, the mining methods of the "isA" and "Consequent" relations between "Intent" nodes are elaborated. 
It is worth noting that the other two relations (i.e., "Consist" and "Has") have been obtained in the "KG Nodes Mining" Section.

\textbf{"isA" Relation:}
Since "isA" is used to organize "Intent" nodes in a hierarchical tree structure, it is challenging to acquire the knowledge that belongs to common sense through data mining.
For instance, it is easy to know that "buy an iPhone13" is a kind of "buy a mobile phone," but difficult for a machine to understand. 
To this end, we propose two different methods described as follows:  \\
1) Lexical Rule-based Method:
This method utilizes the "isA" of the "Product" to build the "isA" relation between "Intent" nodes.
For example, "buy an iPhone13" and "buy a mobile phone" have the same "Function," and it can be obtained from the general knowledge graph that "iPhone13" is a kind of "mobile phone," then the relation of "buy an iPhone13 -isA- buy a mobile phone" can be generated.\\
2) Embedding-based Method:
This method employs the information of the text semantics to avoid the drawbacks of the lexical rule-based method.
Specifically, we first apply StructBERT \cite{StructBERT} pre-trained on Alipay corpus to represent the embedding of the "Product." 
Secondly, we calculate the cosine distance between "Product" nodes and recall the top-K candidates with the closest semantics.
Finally, we sample the triples having an "isA" relation between "Intent" nodes. \par
\textbf{"Consequent" Relation:} Bayesian network \cite{Bayesian} is leveraged from the probability network to mine the "Consequent" relation in AlipayKG.
Specifically, the "Intent" of different time segments is first aggregated as the input of the Bayesian network. 
After learning the Bayesian network structure, it leverages relation inference to obtain numerous pairs of "Intent" nodes. Finally, we can build the "Consequent" relation on pairs of highly correlated and order-sensitive "Intent" nodes.

\begin{figure}[t]
\centering
\includegraphics[width=1\linewidth]{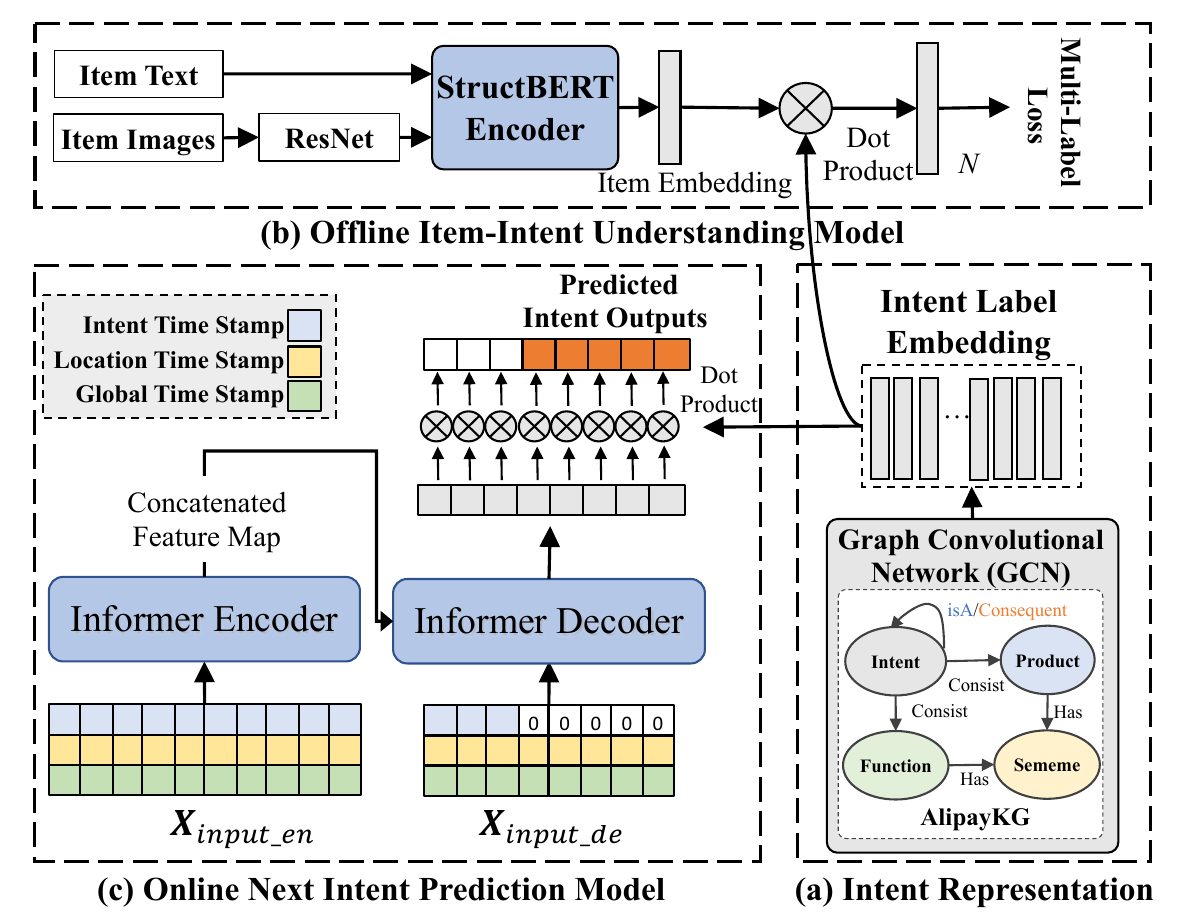}
\caption{
User next intent prediction framework.
 Figure(a): GCN is learned over the AlipayKG to obtain intent label representation, which is applied to predict output intents. 
 Figure(b): Intent label generation of each item via the multi-label classification model. 
 Figure(c): 
 1) The encoder receives massive long sequence inputs (intent, location and global time); 
 2) The decoder receives long sequence inputs, pads the target intents into zero, and instantly predicts output intents (marked orange) in a generative style.}
\label{Next Intent Model}
\end{figure}
\subsection{Next Intent Prediction Framework}
Figure \ref{Next Intent Model} illustrates the next intent prediction framework, which consists of two parts: 1) Offline Item-Intent Understanding Model to label the user interacted items with "Intent," and 2) Online User Next Intent Prediction Model to forecast the next intent of users with low latency and high prediction accuracy.

\subsubsection{Offline Item-Intent Understanding Model}
\label{match-model}
Since user intent is always hidden in the items that users interact with, it is important to establish the Item-Intent relationships, which can be regarded as a matching problem between "Item" and "Intent."
For example, "Starbucks applet" contains various "Intent" such as "order coffee" and "buy coffee beans." 

The overview of the item-intent understanding model is shown in Figure \ref{Next Intent Model}(b). 
Firstly, we follow \citet{MMBT} to unify the multi-source heterogeneous input data. 
Specifically, we adopt Resnet to extract image features and combine them with text features. 
Then, the concatenated features are fed into StructBERT \cite{StructBERT} model to obtain the item representation. 
Besides, intent embedding is generated via graph algorithms such as GCN as shown in \ref{Next Intent Model}(a). 
Finally, we can obtain the intent label scores by matching the learned intent embedding with item representation.

\subsubsection{Online User Next Intent Prediction Model}
Online real-time next intent prediction model needs low latency while guaranteeing high prediction accuracy. 
Hence, an efficient Transformer-based model for long time-series forecasting named Informer \cite{Informer} is adopted in our work.
In this model, the input consists of three parts: the intent timestamp, the location timestamp and the global timestamp (e.g., the time including Minutes, Hours, Weeks and Months when the intent occurs.). 
Moreover, AlipayKG is fused into the model to enhance the prediction accuracy, as shown in Figure \ref{Next Intent Model}(c). 
Additionally, the mined rules (such as "take an internet taxi -consequent- buy movie tickets -consequent- buy snacks") are applied to the post-processing stage (by literal mapping) of the model, which further improves the interpretability of the predicted results.

\subsection{Industrial Deployment}
\label{deploy}
\begin{figure}[t]
\centering
\includegraphics[width=1\linewidth]{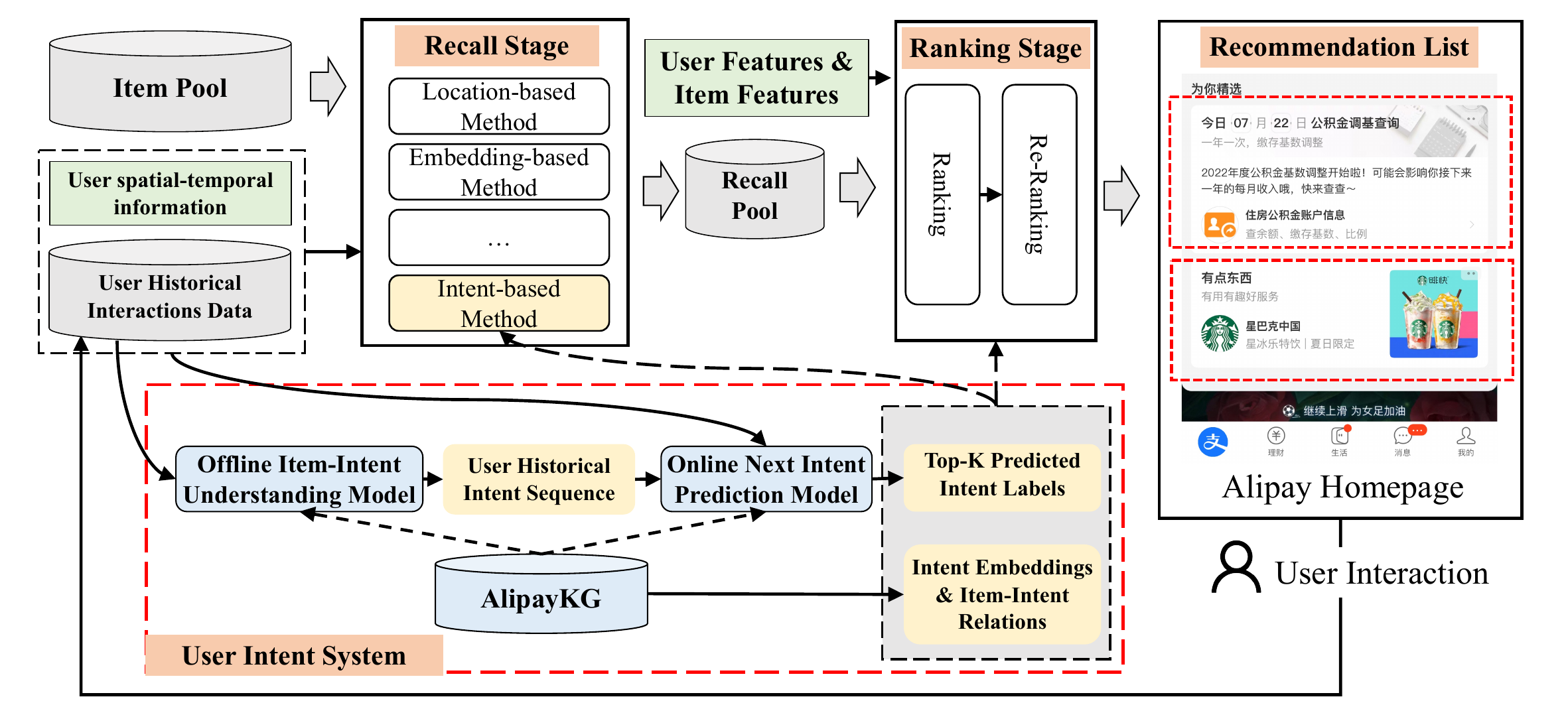}
\caption{
Industrial deployment of User Next Intent Prediction System in the Alipay recommendation engine. 
The recommendation engine contains two stages: the recall stage and the ranking stage. 
The dataflows of recommended items are guided by the grey arrows. The system provides intent embeddings, item-intent relations and top-K predicted intents based on historical information, thereby improving the performance of the recall and ranking stages and providing users with a more in-demand recommendation list.}
\label{Industrial deployment}
\end{figure}

\textbf{The proposed system has now been deployed at Alipay for a few months, to support a variety of business scenarios, achieving better online performance.}
First of all, it can be observed from Figure \ref{Industrial deployment} that the recommendation engine is composed of a recall stage and a ranking stage. 
In the recall stage, a candidate item set (recall pool) is generated by merging results from different recall methods. 
In the ranking stage, those candidates are passed through ranking and re-ranking to output the final recommendation list. 
Secondly, the proposed user intent system is applied to the recommendation engine in the recall and ranking stages.
As shown in Figure \ref{Industrial deployment}, according to history behavior data and current spatial-temporal information, the next intent prediction model can predict the user's top-K intent candidates with the highest probability, which helps bring the intent-based recall method directly into the recall stage. 
Meanwhile, the generated Top-K intent candidates, intent embedding and item-intent relations can contribute to better-personalized modeling of user behaviors in the ranking stage. 
Finally, the whole system is in a positive feedback loop, as shown in Figure \ref{Industrial deployment}.
User Intent System can predict user intent based on user-interacted data, which facilitates better recommendations. 
In return, a better recommendation can provide more real user behavior data to improve the performance of intent understanding. 
Besides, we illustrate the efficacy of the deployment in Section \ref{eval-app}.

\section{Evaluation}
\subsection{Evaluation of AlipayKG} 
 
In AlipayKG, we have accumulated $104K^+$ "Intent," $31K^+$ "Function," $66K^+$ "Product," and $1.9K^+$ "Sememe."
With the item-intent understanding model, we have collected relatively static data, such as $1,316K^+$ Service-Intent triples and $57,852K^+$ Store-Intent triples, and relatively dynamic data, such as 10K-level Coupon-Intent triples and billion-level Bill-Intent triples, etc. 

\subsection{Evaluation of Next Intent Prediction Framework} 

In this Section, we evaluate the proposed intent prediction framework from the following two aspects:\\
\textbf{1) Offline Item-Intent Understanding Model:} We evaluate our matching model on item-intent prediction with $3K^+$ primary intent labels.
In this task, we choose the multi-label classification model \cite{Asymmetric} based on StructBERT\cite{StructBERT} Encode as the \textbf{baseline}.
Compared with the baseline method, our model introduced in Section \ref{match-model} brings an increase of 3.08\% and reaches 90.64\% in micro-F1.\\
\textbf{2) Online Next Intent Prediction Model:} We evaluate our next-intent prediction model on $30K$ sampled user historical behavior data. 
To restore online scenarios, we only predict the user's next intent at a specific time and location. 
Experimental results show that \textbf{the baseline method} based on the Informer \cite{Informer} model achieves 50.2\% and 83.1\% in Recall@1 and Recall@10, respectively. After fusing the Intent Graph Embedding shown in Figure \ref{Next Intent Model}(b), the metrics of Recall@1 and Recall@10 increase by 3.1\% and 2.2\%, respectively.


\subsection{Evaluation of Downstream Applications}
\label{eval-app}
In this section, we further evaluate whether the user next intent prediction system can improve the downstream tasks' performance at Alipay.
We divide users into buckets for the online A/B test.\\
\textbf{1) Home Recommendation:} Home recommendation is one of the essential business scenarios in which our system helps to discover user interests in real-time, as shown in Section \ref{deploy}. 
Online experiments demonstrate that the recommendation performance with the user intent system improves by 1.61\% in CTR (Click-Through-Rate). Such a relative increase proves that our system can enhance the recommendation with the real-time predicted user intent.\\
\textbf{2) Transaction Risk Management:} To create a secure payment environment, the potential risks (e.g., embezzlement and money laundering) of each transaction should be estimated to determine whether it is invalid, which consumes a huge amount of computation. To reduce the cost, we treat users' consumption intent as an important transaction feature to discover low-risk transactions. By leveraging the credible transaction identification based on the proposed user intent system, the coverage rate of low-risk transactions is relatively increased by 100\%. \\

\section{Conclusion and Future Work}
In this work, we present the user intent system and demonstrate its effectiveness in downstream applications deployed in Alipay. 
In the future, we will continually maintain the AlipayKG to cover more business data and applications, and hopefully, it can benefit more downstream tasks in digital life.
Furthermore, we will make efforts in the direction of interpretable reasoning for better user intent prediction.

\begin{acks}
This work was supported by the National Natural Science Foundation of China (No.62206246), Zhejiang Provincial Natural Science Foundation of China (No. LGG22F030011), Ningbo Natural Science Foundation (2021J190), and Yongjiang Talent Introduction Programme (2021A-156-G).
\end{acks}

\end{CJK}

\bibliographystyle{ACM-Reference-Format}
\bibliography{custom}

\end{document}